\pdfoutput=1

\documentclass[11pt]{article}

\usepackage{EMNLP2022}

\usepackage{times}
\usepackage{latexsym}

\usepackage[T1]{fontenc}

\usepackage[utf8]{inputenc}

\usepackage{microtype}

\usepackage{inconsolata}

\usepackage{amsfonts}
\usepackage{algpseudocode}
\usepackage{csquotes}
\usepackage{graphicx}
\usepackage{enumitem}
\usepackage{blindtext}
\usepackage{multicol}
\usepackage{CJKutf8}
\usepackage{cancel}

\usepackage[utf8]{inputenc}
\usepackage[version=4]{mhchem}

\DeclareMathOperator*{\argmin}{argmin}

\title{Generative Models as a Complex Systems Science:\\ How can we make sense of large language model behavior?}

\author{Ari Holtzman$^1$\and Peter West \and Luke Zettlemoyer \\
    University of Washington \\ \texttt{\{ahai,pawest,lsz\}@cs.washington.edu} \\
 }

\begin{document}
\maketitle

\textit{}\newcommand{\ari}[1]{\textcolor{purple}{ari: #1}}
\newcommand{\peter}[1]{\textcolor{red}{peter: #1}}
\newcommand{\luke}[1]{\textcolor{red}{luke: #1}}
\newcommand{\alisa}[1]{\textcolor{red}{alisa: #1}}
\newcommand{\tim}[1]{\textcolor{red}{tim: #1}}
\newcommand{\gabriel}[1]{\textcolor{red}{gabriel: #1}}
\newcommand{\jared}[1]{\textcolor{red}{jared: #1}}
\newcommand{\dallas}[1]{\textcolor{red}{dallas: #1}}
\newcommand{\sewon}[1]{\textcolor{red}{sewon: #1}}
\newcommand{\julian}[1]{\textcolor{red}{julian: #1}} 
\newcommand{\ian}[1]{\textcolor{red}{ian: #1}}

\newcommand{\newmodel}[0]{Newformer\xspace}
\newcommand{\goal}[0]{taxonomy\xspace}
\newcommand{\lm}[0]{LM\xspace}
\newcommand{\blackbox}[0]{black-box\xspace}
\newcommand{\whitebox}[0]{white-box\xspace}
\newcommand{\inferencedata}[0]{inference data\xspace}

\newcommand{\interalia}[1]{\cite[][\textit{inter alia}]{#1}}

\newcommand{\model}[0]{\mathcal{M}}
\newcommand{\task}[0]{\mathcal{T}}
\newcommand{\inpset}[0]{\mathcal{X}}
\newcommand{\outset}[0]{\mathcal{Y}}
\newcommand{\randset}[0]{\mathcal{R}}
\newcommand{\featset}[0]{\mathcal{F}}
\newcommand{\behavior}[0]{\mathcal{B}}
\newcommand{\inpvar}[0]{\mathbf{X}}
\newcommand{\outvar}[0]{\mathbf{Y}}
\newcommand{\randvar}[0]{\mathbf{R}}
\newcommand{\behvar}[0]{\mathbf{B}}
\newcommand{\mi}[0]{I}
\newcommand{\ent}[0]{H}
\newcommand{\dist}[0]{\mathcal{D}}
\newcommand{\code}[0]{\mathcal{C}}
\newcommand{\objset}[0]{\mathcal{O}}
\newcommand{\strset}[0]{\mathcal{S}}
\newcommand{\exinpvar}[0]{\mathb{E}}
\newcommand{\exoutvar}[0]{\mathb{A}}

\begin{figure*}[t]
    \centering
    \includegraphics[width=\linewidth]{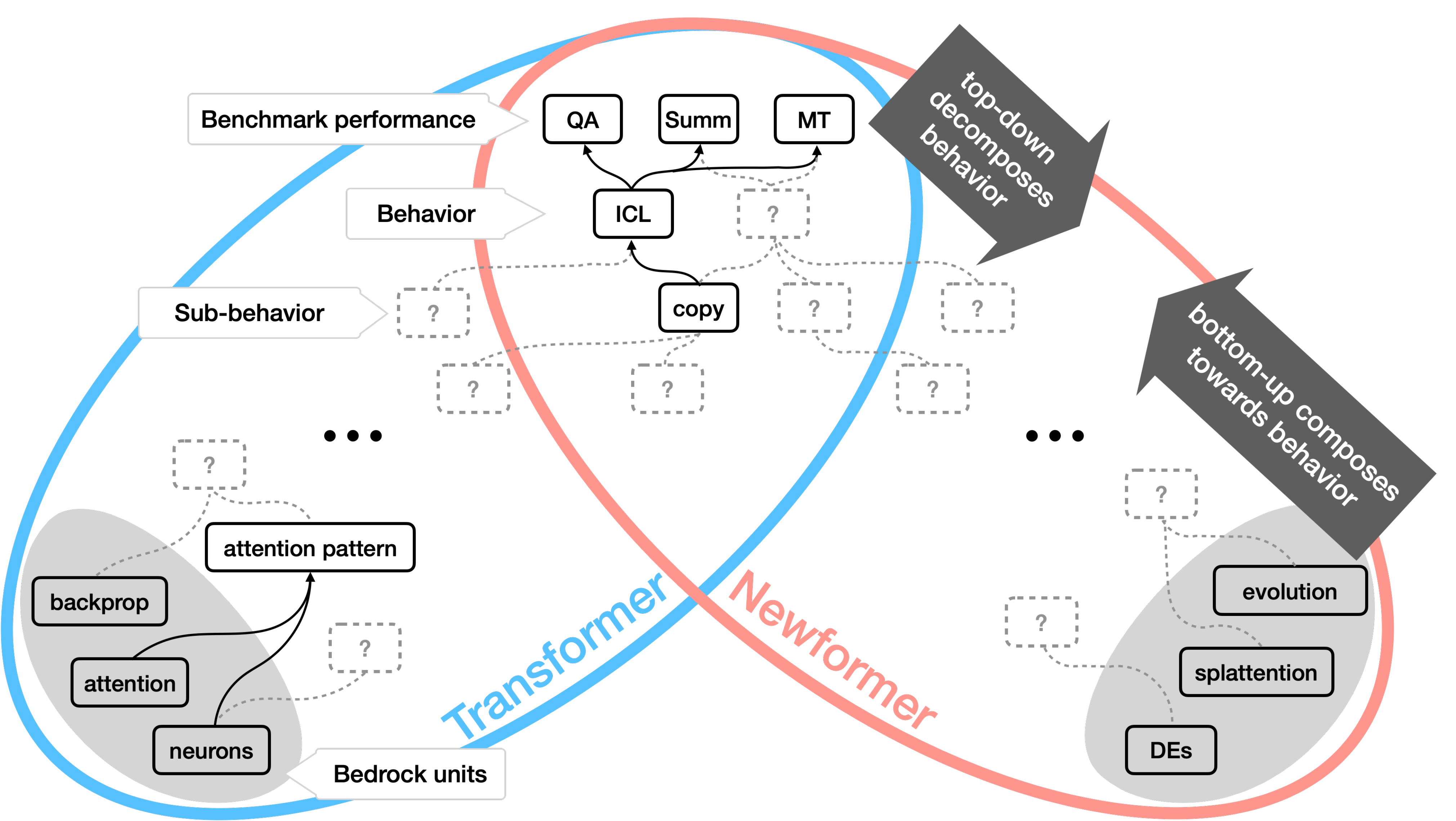}
    \caption{To explain why learned models self-organize the way they do from the bottom-up, it is useful to have top-down hierarchy of partially decomposed behaviors, to guide hypotheses with functionality we know the overall model has.
    While networks are composed of bedrock units for which we have a perfect understanding by construction (e.g. neurons), 
    most emergent aspects of these systems are still undefined and undiscovered (represented as ``?''). 
    }
    \label{fig:hierarchy}
\end{figure*} 
\begin{abstract}

\renewcommand\thefootnote{1}\footnotetext{Incoming faculty at University of Chicago}

\renewcommand*{\thefootnote}{\arabic{footnote}}
\setcounter{footnote}{0}

Coaxing out desired behavior from pretrained models, while avoiding undesirable ones, has redefined NLP and is reshaping how we interact with computers. What was once a scientific engineering discipline—in which building blocks are stacked one on top of the other—is arguably already a complex systems science—in which \textit{emergent behaviors} are sought out to support previously unimagined use cases. 

Despite the ever increasing number of benchmarks that measure \textit{task performance}, we lack explanations of what \textit{behaviors} language models exhibit that allow them to complete these tasks in the first place. 
We argue for a systematic effort to decompose language model behavior into categories that explain cross-task performance, to guide mechanistic explanations and help future-proof analytic research.
\end{abstract}
\section{The \newmodel: \hspace{120pt} A Thought Experiment}
\label{sec:newmodel}

Consider the following thought experiment:
\begin{equation*}
\tag{A}\label{te:newformer}
   \parbox{\dimexpr\linewidth-4em}{
    \strut
    Tomorrow, researchers at an industry lab publicly release a new kind of pretrained model: the \newmodel. It has a completely different architecture than the Transformer (no attention, non-differentiable components, etc.), that outperforms all pretrained Transformers on the vast majority of benchmarks. Independent labs quickly verify that these results are sound, even on just-released benchmarks. While the composition of the training data is public, it is so expensive to train that no lab can afford to replicate it, even the one that produced it. Scaled-down versions do not exhibit the same performance or interesting behaviors as the original model.
    \strut
  }
\end{equation*}

\subsection{How should we study the \newmodel?}
\label{ssec:sum}

Identifying high-level behaviors a model does or does not share with older models can steer us toward lower-level mechanisms it uses to solve tasks (\S\ref{ssec:top-down-guides}, Figure~\ref{fig:hierarchy}). Interpretation techniques that rely on low-level details are model specific (\S\ref{ssec:lstms}) and often abandoned as the field changes. The \newmodel is fictional, but it can help us reconceptualize the goals and methods of generative model research in light of the new landscape (\S\ref{ssec:are-we-there-yet}).

\begin{figure*}[t]
    \centering

    \includegraphics[width=\linewidth]{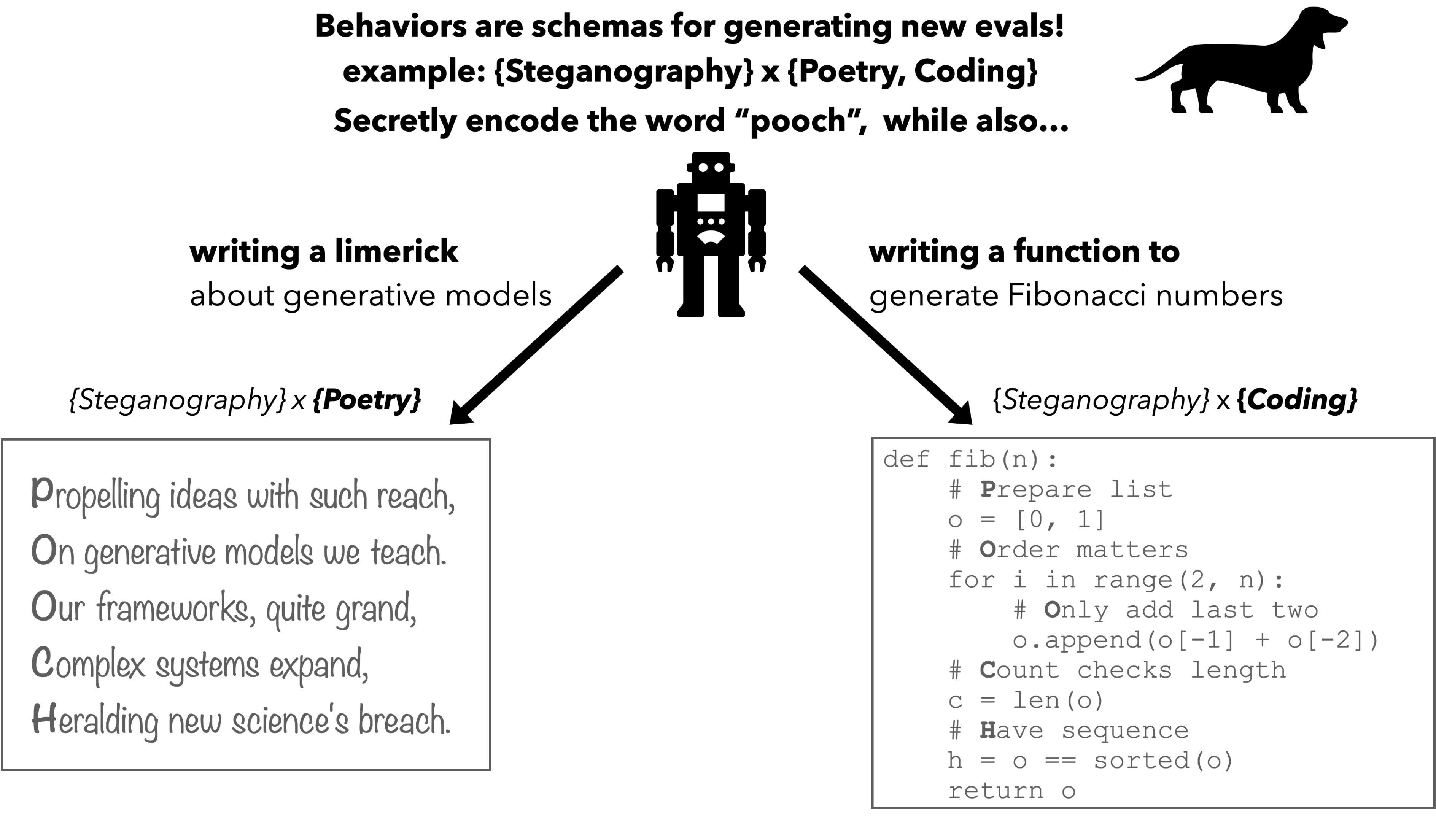}

    \caption{An example of how behaviors can be used to create new evaluations. These examples were generated from GPT-4, but required significant human curation, suggesting that Thought~Experiment~\ref{te:schemas} has not yet occurred.}
    \label{fig:steganography}
\end{figure*}

How should we factorize model behavior into understandable and explanatory categories? (\S\ref{section:behavior}, Figure~\ref{fig:hourglass}) We present a formalism for describing behavior (\S\ref{ssec:formal}), noting that this corresponds to a \textit{metamodel} that predicts aspects of a primary model (Figure~\ref{fig:meta-model-generalization}).
Benchmarks help us measure \textit{performance}, but rarely \textit{discover behavior} (\S\ref{ssec:discover}) or \textit{characterize} it (\S\ref{ssec:characterize}). Instead, discovered behaviors motivate new benchmarks (\S\ref{ssec:building-blocks}, Figure~\ref{fig:steganography}).

Generative models qualify as \textit{complex systems} (\S\ref{sec:complex}), due to their \textit{emergent behaviors} (\S\ref{ssec:what-is}, Figure~\ref{fig:micro-to-macro}), which are more often \textit{discovered} than engineered (\S\ref{ssec:design-vs-discover}). 
A lack of clarity on \textit{what} models do holds us back, as if we were studying organic chemistry without knowledge of biology (\S\ref{ssec:neuronal}). This issue remains even when proprietary models are released (\S\ref{ssec:access}), as the problem lies in our lack of behavioral vocabulary;  investigating possible mappings between training data and generated data can help us establish new behavioral categories (\S\ref{ssec:maac}).

\begin{figure*}[t]
    \centering
    \includegraphics[width=\linewidth]{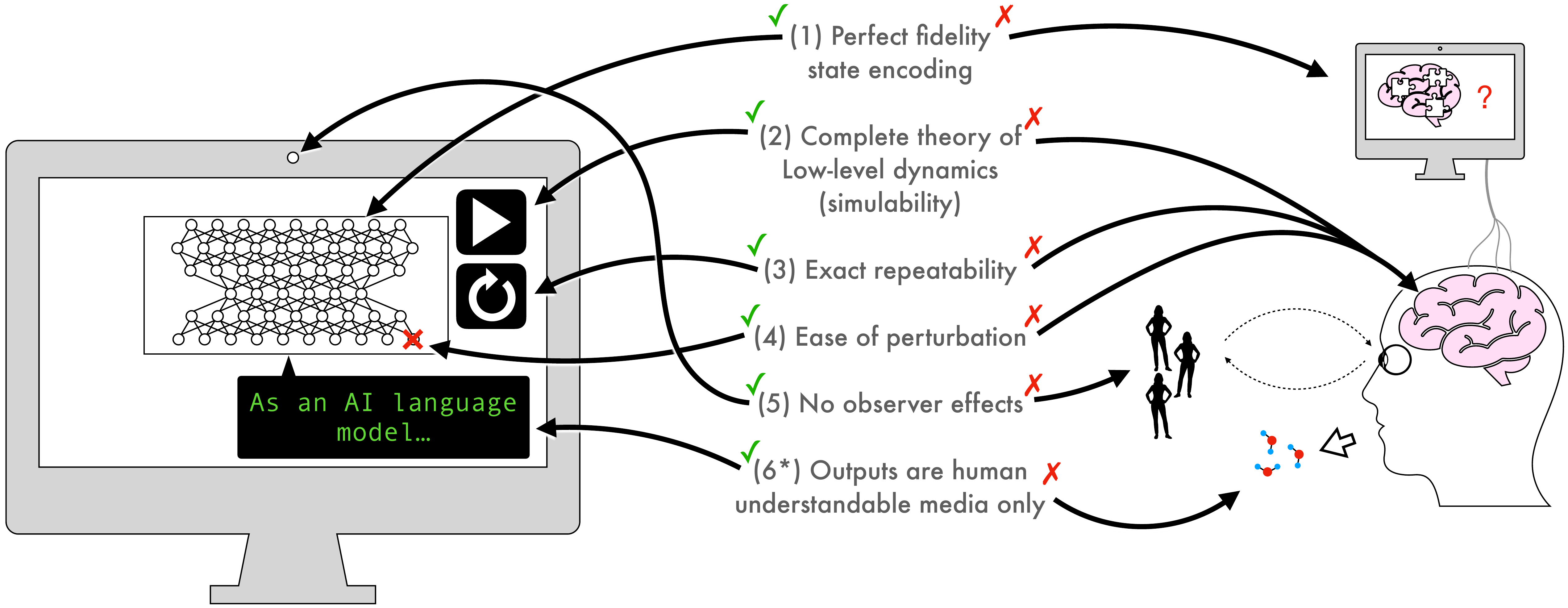}
    \caption{  Visual representation of the advantages generative model researchers have over researchers that study the main other media generating system on earth: human beings.}
    \label{fig:advantages}
\end{figure*}

Despite the challenges, generative models are \textit{easier} to study than many naturally arising complex systems (\S\ref{sec:diff}), because they are simulable by construction (\S\ref{ssec:two-kinds}). In contrast to physical phenomena, we can easily conduct a wide range of storable, repeatable experiments without observer effects (\S\ref{ssec:advantages}, Figure~\ref{fig:advantages}). We do, however, rely on the availability of open-source models (\S\ref{ssec:opensource}). 

We conclude (\S\ref{sec:conclusion}) with an argument for increased focus on the foundational ``what are models doing?'' to guide the classic ``why are models doing that?''

\subsection{Top-down behavioral \goal guides \hspace{200pt} bottom-up mechanistic explanation}
\label{ssec:top-down-guides}

The \newmodel is a completely opaque result when considering benchmarks alone; it is simply better at doing what we want it to do than Transformer models \cite{vaswani2017attention} were before. However, as shown in Figure~\ref{fig:hierarchy}, a hierarchical \goal of \lm behavior can guide our investigation of the \newmodel, leading to questions such as:
\begin{enumerate}
    \item What behaviors do the Transformer models and the \newmodel model share, e.g., does the \newmodel also repeat phrases more often than as seen in the training data?
    \item Do they exhibit similar behaviors in the same contexts, e.g., does the \newmodel need fewer  input-output demonstrations to exhibit in-context learning at peak performance?
    \item Do high-level behaviors decompose into the same lower-level behaviors or does the \newmodel use different mechanisms to express them, e.g., when the \newmodel is used for paraphrasing does it also tend to exactly copy the input? 
\end{enumerate}
Without such behavioral categories, we risk investigating the wrong direction when we try to interpret models, because we do not know what phenomena we are trying to explain in the first place.

Observed behavior can tell us where to look for bottom-up explanations. \citet{al2019character} observed emergent copying behavior in Transformer Language Models (\lm{}s), paving the way for the discovery of \textit{copying heads} that make copying possible. Characterizing copying heads led to the discovery of \textit{induction heads} \cite{elhage2021mathematical, olsson2022icl}: Transformer heads that are capable of copying abstract representational patterns in previous layers and appear to be responsible for in-context learning. \citet{olsson2022icl} show that induction heads exhibit a variety of pattern matching behaviors that are still not fully catalogued.

Attempting to explain neural networks bottom-up without being guided by behavior can make it difficult to interpret results.
For example, many works that identify anisotropy in the embedding spaces of large \lm{}s diagnose this as a deficiency, and attempt to fix it \cite{wang2020improving, ethayarajh2019contextual, gao2019representation}. However, recent work suggests that this anisotropic property may not actually limit expressivity \cite{bis2021too}, may be a result of the transformer architecture specifically \cite{godey2023anisotropy}, and may actually be helpful for language models \cite{rudman2023stable}.

\subsection{The Transformer is the old \newmodel}
\label{ssec:lstms}

A moderate \newmodel event has occurred at least once before with \lm{}s: the switch from Recurrent Neural Networks (RNNs) to Transformers. Despite many partial explanations \cite{lakretz-etal-2019-emergence, Olah2015-kq, hochreiter1997long}, we still lack an explanatory theory of how LSTM \cite{hochreiter1997long} \lm{}s such as ELMo \cite{peters-etal-2018-deep} worked—what behaviors they could and could not capture, how these composed, etc.—even as they were replaced by models like BERT \cite{devlin2019bert} with similar use cases, but completely different architectural details. This does not bode well for the introduction of something like the \newmodel which is significantly farther from the Transformer than the Transformer is from the LSTM.

On their own, bottom-up methods do not transfer well to new systems: analysis techniques that relied on mutated state and gating in RNNs, such as visualizing gating mechanisms \cite{Karpathy2015-xl}, are not applicable to Transformers. Interpretation methods for Transformer models \cite{weiss2021thinking, rogers2020primer}, such as those that use attention, are unlikely to transfer over to the \newmodel which breaks many previously immutable assumptions. 

This suggests the value in doing more interpretation work that
treats models like \textit{\blackbox}es, as if we do not have access to their internal mechanisms. There is growing interest in looking at NLP systems as \blackbox{}es \cite{bastings2022proceedings, ribeiro2020beyond, blackboxnlp-2019-blackboxnlp}, though much of this work still uses intermediate outputs—such as embeddings—rather than directly analyzing behavior in the output space models are trained to fit. 
Truly \blackbox methods can help insulate our analysis from change, giving us an anchor point that will always be testable on models that use the same modality (e.g., text, speech, images). \citet{belinkov2017synthetic} show that neural machine translation systems are brittle to both natural spelling errors and synthetic   character-level noise. This observation can be extended to ask: Is the \newmodel robust to the same kinds of noise? Up to what threshold? Does the noise appear to be localized in the brittleness of tokenization, as was the case for Transformer-based systems \cite{provilkov2020bpe}? Developing a rich inventory of such tests would give us a universal scaffolding for analyzing any \newmodel the moment it is discovered.

\subsection{Are we there yet?}
\label{ssec:are-we-there-yet}

Deciding whether a \newmodel-like event has \textit{truly} happened is an unresolvable question. New models are always partially derivative, and new (possibly artificial) axes can always be invented where they are worse or better \cite{Wolpert1997-nn}. Yet three years on it is still infeasible for most labs to train a GPT-3 \cite{brown2020language} level \lm{}, costing approximately half a million dollars in compute alone for private companies—with engineering teams—to produce a similar model \cite{Abhinav_Venigalla2022-sk}. Thus it seems that the gap for training is only growing wider as ChatGPT \cite{openai2022chatgpt} and GPT-4 \cite{OpenAI2023gpt4} become commonplace in research \cite{yang2023harnessing, zhang2023one}, production \cite{eloundou2023gpts, george2023review, ray2023chatgpt},  and even model evaluation \cite{liu2023gpteval, Zheng2023-fk}.

Unlike the \newmodel these models were never released, are frequently deprecated \cite{openai2023avail}, change from day to day \cite{Perry2023-df, Southern_undated-kp}, and are know to be unstable over theoretically deterministic queries \cite{Deng_undated-cl}. Yet, the open source community has caught-up quickly \interalia{Alizadeh2023-pd, Mukherjee2023-cw, gunasekar2023textbooks}
helped by industry labs' open-sourcing efforts \interalia{Touvron2023-ml, falcon40b, stablelm}, and new finetuning tehchniques \cite{alpaca, vicunaTeam-ja, dettmers2023qlora}.

However, the question still remains: how should we explain models not everyone can train? Models that are so arduous, slow, and expensive to train that we will likely never ablate all the necessary variables needed to study them properly? 

This leaves us with mere behavior. We generally think of there as being two different kinds of behavior: the neural behavior of different activations in models and the ``output'' of the model in the form of human media (e.g., text, images, videos, etc.). Most methods of explaining models focus on the former: trying to explain why neural activations cluster into certain patterns and trying to understand what those patterns mean about the output.

 We argue that not enough attention has been given to formalizing the latter: \textit{what} models are doing in the first place, in terms of regularities in their outputs. Without such a formalization, bottom-up methods will have a much harder time deciding what precisely to explain, and what is simply noise.

\clearpage

\section{The Behavioral Bottleneck}
\label{section:behavior}
\begin{figure*}[hp]
    \centering

    \includegraphics[width=\linewidth]{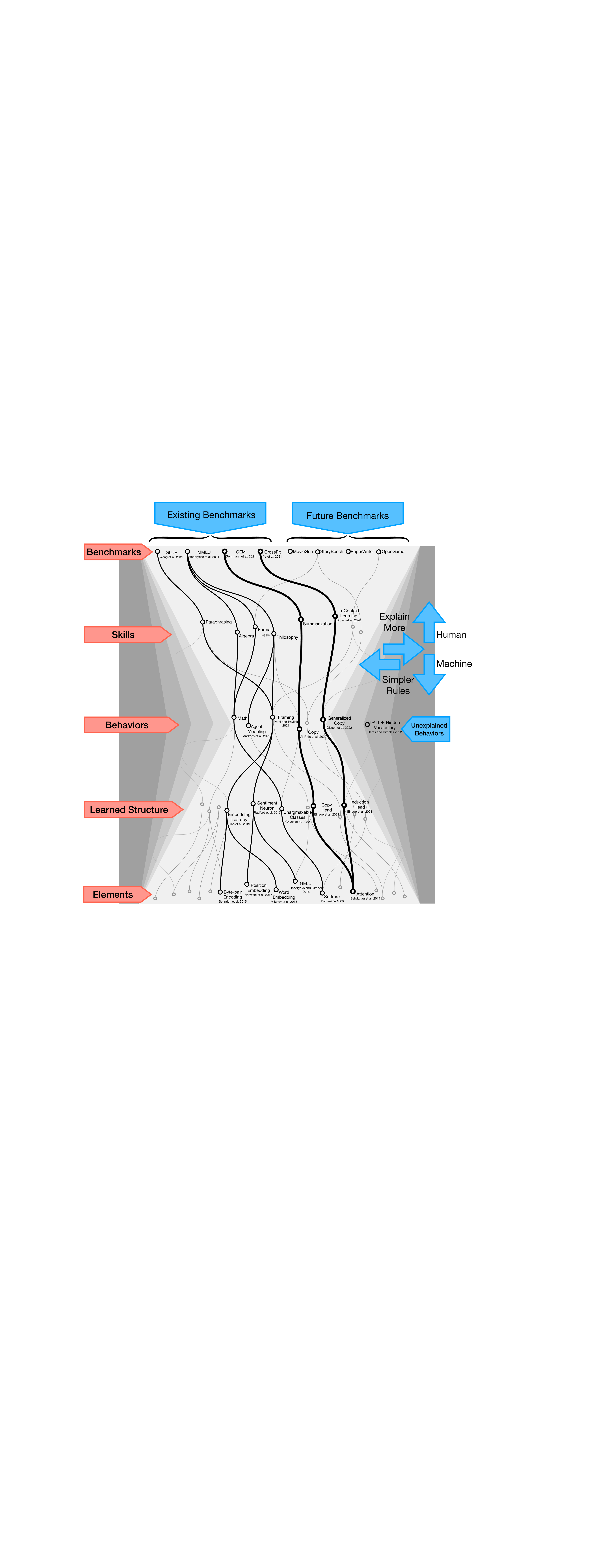}
    
    \caption{A visual representation of different aspects of models, shown from the basic elements of models on the bottom up to the benchmarks we are attempting to solve. Nodes represent invented and discovered aspects of models. The highlighted subgraph captures the concepts that we might want to use for understanding the phenomenon of ``copying'' in Transformers, when models generate sequences that appear in their local context window, a behavior that serves as a running example in this paper. \\ \\ We might start out by noticing that Transformer models have higher scores on GEM \cite{gehrmann2021gem} (\textbf{Benchmark}), especially on summarization-like tasks (\textbf{Skill}). Inspecting the data generated by the models of interest, we might notice one of the qualitative differences separating Transformer models from other models is the ability to correctly use novel entities \cite{al2019character} (\textbf{Behavior}). 
    We might ask why this is, embarking on an empirical study of when networks develop the ability to copy, as \citet{elhage2021mathematical} did, discovering specific attention heads served as \textit{copying heads} (\textbf{Learned Structure} supported by certain \textbf{Elements}). This led to other discoveries such as \textit{induction heads} \cite{elhage2021mathematical, olsson2022icl} (\textbf{Learned Structure}), which were found to perform a kind of \textit{generalized copying} that supports inference-time pattern recognition (\textbf{Behavior}), e.g., for In-Context Learning \cite{brown2020language} (\textbf{Skill}), leading to better results on fewshot benchmarks such as CrossFit \cite{ye-etal-2021-crossfit} (\textbf{Benchmark}). Research can proceed by observing high-level behavioral regularities, explaining them via  the tendencies of the model, and using this to achieve clarity about other observed behaviors.
    \nocite{wang2018glue}
    \nocite{hendrycks2020measuring}
    \nocite{gehrmann2021gem}
    \nocite{ye-etal-2021-crossfit}
    \nocite{brown2020language}
    \nocite{andreas2022language}
    \nocite{patel-pavlick-2021-stated}
    \nocite{al2019character}
    \nocite{olsson2022icl}
    \nocite{daras2022discovering}
    \nocite{gao2019representation}
    \nocite{radford2017learning}
    \nocite{grivas-etal-2022-low}
    \nocite{elhage2021mathematical}
    \nocite{sennrich2016neural}
    \nocite{vaswani2017attention}
    \nocite{mikolov2013distributed}
    \nocite{hendrycks2016gaussian}
    \nocite{boltzmann1868studien}
    \nocite{bahdanau2014neural}
}
    \label{fig:hourglass}
\end{figure*}

How do we avoid proposing a new explanation
for every exhibited difference? Surely we do not believe that we need a benchmark for every prompt that elicits slightly different behavior from a generative model? One solution is to propose many possible mechanisms, but make it an explicit research agenda to discover \textit{the most parsimonious explanation}, a concept visualized in Figure~\ref{fig:hourglass}. In other words, we want to be able to predict the aspects of text we care about (e.g. factuality) with the simplest rules possible.
We briefly formalize this concept in \S\ref{ssec:formal}, but the bulk of this paper concerns the \textit{need} for this new research focus and the perspective it yields.

Thousands of papers observe behavioral tendencies in models, such as the ability of a pretrained Transformer to copy from the input context \cite{elhage2021mathematical, al2019character}, which we will adopt as a running example. To understand models better, we must rigorously describe (1) what \textit{aspect} of generative behavior a given mechanism predicts (e.g. repetition, copying from the training set, etc.) and (2) how much of the \textit{information} in the output space of the model such predictions explain (since most will not predict 100\% of what a model emits).

Figure~\ref{fig:hourglass} serves as a visual map of how we might explain models via behavior. On the top level we have a huge diversity of benchmarks that currently exist, and the even larger number that may one day exist. On the bottom we have the mathematical abstraction that describes the space of all possible models. Clearly both of these represent many more possibilities than is useful as an explanation or than is \textit{necessary} to explain specific facets of model behavior. The intermediate levels, then, deal with simplified metamodels, i.e., models of the underlying generative model that are less explanatory, but still allow us to interpret or theorize around models.

\subsection{A working definition of ``behavior''}
\label{ssec:formal}

\citet{fong2017interpretable} state that: ``An explanation is a rule that predicts the response of a black box $f$ to certain inputs.'' We think of a \textit{behavior} as an explanation of limited aspects of a model, a concept we briefly formalize. We make reference to this formalization sparsely throughout the rest of the paper, as the argument can be understood without it, and we stress that the problem we are facing is more fundamental than a missing formalism.

\newpage
Given a generative model from one input medium $\inpset$ (e.g., strings composed of at most 2048 tokens) and a source of randomness $\randset$ to an output medium $\outset$ (e.g., 512x512 pixel images):
\begin{align}
    \label{eq:model}
    \model: \inpset \times \randset \to \outset
\end{align}
we can define a behavior as a function from the same input medium to a feature set $\featset$: 
\begin{align}
\behavior: \inpset \to \featset
\end{align}
For instance, $\model$ may be a general purpose text-to-image model trained on scraped data, while $\behavior$ may map a string $x \in \inpset$ to a probability that an image $\model(x)$,  contains at least one dog. Or $\inpset$ and $\outset$ may both be Unicode strings, in the case of an \lm, with $\behavior$ being a binary prediction as to whether $\model(x)$ will eventually get caught in a repetition loop \cite{holtzman2020curious}. 

Our goal in proposing behaviors is to \textit{explain} the underlying model using rules that capture model tendencies. Behaviors are explanatory to the extent that they give us information about the application of the model $\model$ under distributions $\dist_\inpset$ and $\dist_\randset$ over $\inpset$ and $\randset$, which we collectively refer to as $\dist$ for brevity. We can formalize the notion of ``giving us information about the application of the model'' through the mutual information: 
\begin{align}
\mi_\dist(\model(\inpvar, \randvar);\behavior(\inpvar)) = \\
\ent_\dist(\model(\inpvar,\randvar)) - \ent_\dist(\model(\inpvar, \randvar)|\behavior(\inpvar))
\end{align}
where $\inpvar$ and $\randvar$ are random variables drawn from $\inpset$ and $\randset$ according to $\dist_\inpset$ and $\dist_\randset$, and $H$ is the entropy: 
$H(\mathbf{Y}) = \mathbb{E}_\dist \left [ -\log p_\dist(\mathbf{Y}) \right ]$ for a random variable $\mathbf{Y}$. The mutual information is a direct measure of \textit{how many bits of information we learn about one variable given another}, so this formulation directly tells us how much a behavior reveals about expected model output.

We call setting $\dist_\inpset$ to be uniform over $\inpset$ the ``mechanistic distribution.'' In this case, the mutual information is unrelated to the expected distribution of inputs in the wild, but is instead representative of how well we can model \textit{any} input to $\model$. For instance, explaining an \lm{} under the mechanistic distribution would require a behavior that predicts aspects of the \lm{}'s output accurately even for long strings of gibberish. This may be difficult, since we often use human linguistic features to make predictions about model outputs, but such behaviors are closer to the notion of mechanistic interpretability that tries to fully reverse engineer the model being studied \cite{Olah2022-hw}.

If $\mathcal{M}$ ignores its source of randomness, i.e., $\mi(\model(\inpvar, \randvar);\randvar) = 0$—as is the case for deterministic models such as a greedy-decoded \lm{}—then the most explanatory behavior is simply $\behavior = \model(\inpvar,r)$ for any $r \in \randset$. This is a degenerate behavior, in that it is very explanatory, but has not brought us any closer to explaining  $\model$. Therefore, we would like behaviors that are not just very high mutual information with the model, but also \textit{point to predictable regularities} in $\model$, especially in a way that allows us to build up new hypotheses about it. Much has been written about what makes an explanation useful \interalia{lipton2018mythos, jacovi2020towards, chen2023machine}, and reviewing these desiderata is out of scope for this paper.

The mutual information $\mi_\dist(\model(\inpvar, \randvar);\behavior(\inpvar))$ can also be viewed as \textit{how much a behavior allows us to compress the output of a model} under a distribution $\dist$, e.g., a distribution of articles for a summarization task (and a random number generator $\randvar$). This is because any bits of information revealed by one variable, can be used to compress the other, under a proper coding scheme \cite{cover1999elements}. 

The concept of Minimum Description Length (MDL) has been used as an information theoretic criterion for finding good hypotheses \cite{grunwald2007minimum}. Essentially, it suggests an extension to Occam's razor \cite{Barry2014-wr}: that we should favor explanations that are simple to describe and explain the object under study the most. We can formalize this notion for behaviors, via an encoding scheme $C$ that represents behaviors $\behavior$ and outputs $y \in \outset$ as binary strings of variable (but finite) length $s \in \strset$, where $|s|$ is the length of a string $s$. A naïve MDL objective would then be:

\begin{align}
    \label{eq:mdl-orig}
    \argmin_{\behavior} |\code(\behavior)| + \sum_{x \in \inpset}\mathbb{E} _{\dist _\randset} \left [ \left | \code (\model(x, \randvar)|\behavior(\inpvar)) \right | \right ] 
\end{align}

\begin{figure}[t]
    \centering

    \includegraphics[width=\linewidth]{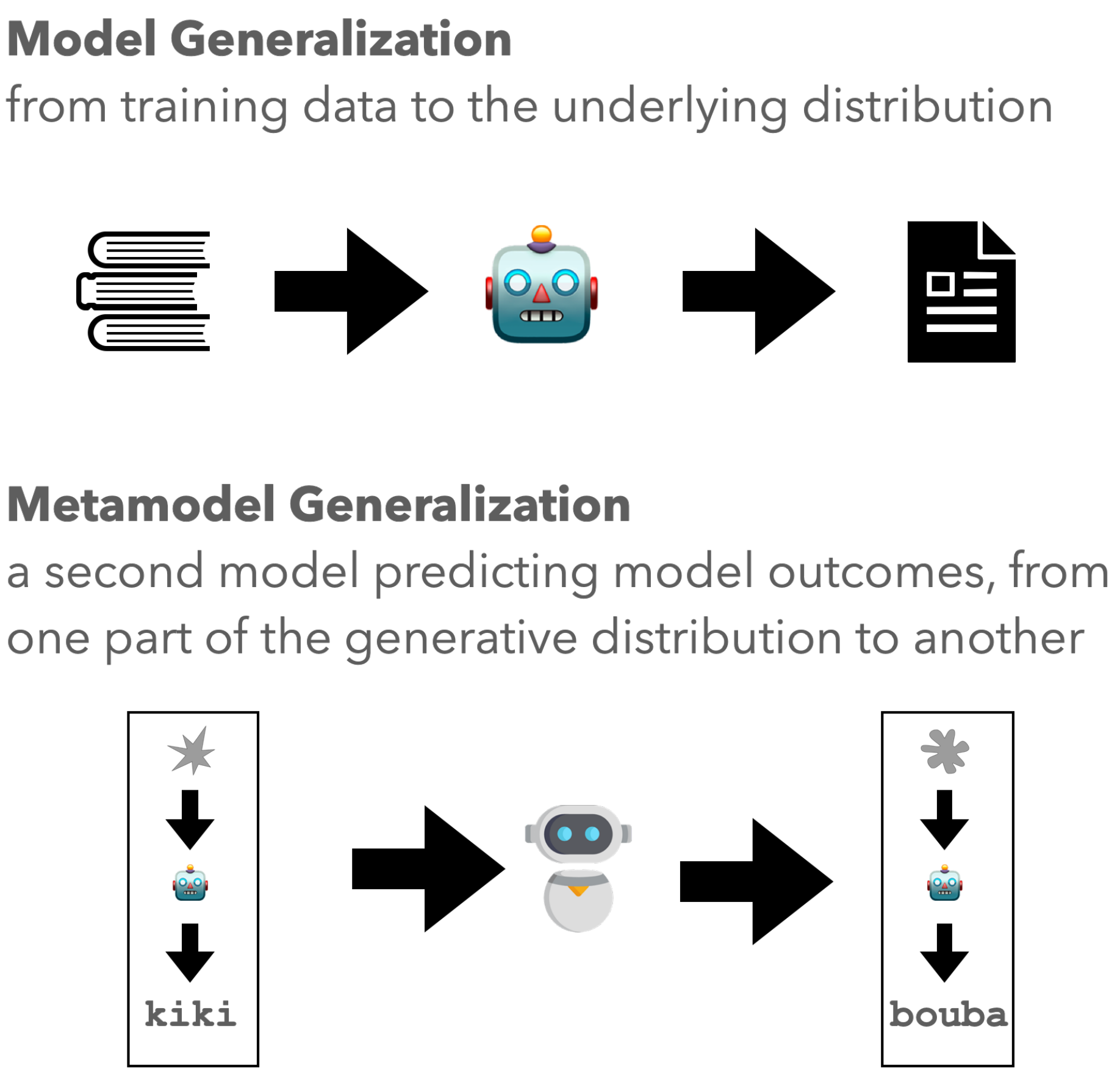}

    \caption{If we cannot train models easily, but those models are sufficiently general and useful, we can predict what models can and can't do, rather than what a model trained differently would do. The kiki/bouba categorization is a cross-culturally robust linguistic-conceptual mapping in humans \cite{Cwiek2022-al}.}
    \label{fig:meta-model-generalization}
\end{figure}

However, this would not suit our general objective: we do not necessarily wish to encode \textit{all possible data a model could produce}, especially since most models have huge output spaces of largely low probability density. Instead, we would like to quantify the information behaviors can save us under $\dist_\inpset$.

To capture the idea ``how much space does $\behavior$ save us under $\dist_\inpset$ we can use:
\begin{align}
    \label{eq:mdl}
    \argmin_{\behavior} |\code(\behavior)| + \alpha \ent_\dist(\model(\inpvar, \randvar)|\behavior(\inpvar))
\end{align}
where we replace the second term with the conditional entropy $\ent_\dist$, since this describes the minimum number of bits that could be used to represent the information encoded \cite{cover1999elements}.
This can be interpreted as, ``we would like behaviors that on average, save us more space in terms of encoding the possible outcomes of a model than they take to describe.'' $\alpha$ allows us to trade-off how much we weight the representation of the behavior vs. the outputs of a model, where larger values of $\alpha$ may be appropriate if we are dealing with a many outputs, making the bits saved by way of conditioning on the behavior more pertinent.

Overall, we seek to find behaviors that are both explanatory and simple to describe. We can think of this as attempting to find a \textit{metamodel}: models that are designed or trained to predict another model's behavior \cite{barton2006metamodel}, as illustrated in Figure~\ref{fig:meta-model-generalization}. 
This suggests we want to find  \textit{behaviors that transfer over different contexts} so we can predict where models will be useful and where they will break down. 

\clearpage

\subsection{Can benchmarks discover new behavior?}
\label{ssec:discover}

In general, discrepancies in performance between benchmarks can \textit{hint} at potentially new behavior, but they cannot discover behavior we have not yet observed. Given the diversity of NLP benchmarks,
it is likely that the \newmodel (\S\ref{sec:newmodel}) will perform drastically different on certain pairs of benchmarks we believe to be related, e.g., the same task in two different domains. This is a useful signal for where to inspect behavior, but benchmarks alone cannot reveal new abilities, underlying mechanisms, or shortcut heuristics the \newmodel is relying on that \textit{cause} a discrepancy in results and what else its effects are.

For example, it is very difficult to imagine how \textit{prompting} \cite{radford2019language, liu2021pre} could have been discovered via benchmarking. 
Finetuning a generative model, such as GPT-2 \cite{radford2019better}, and doing well or badly at any number of benchmarks could not have revealed a model can be prompted with text that matched training data patterns, to elicit behavior such as summary generation via the string ``TL;DR'' 
or translation through formatting such as ``French sentence: <source>$\backslash\backslash$English sentence:''. These discoveries are a result of inspecting the generative \textit{behavior} of GPT-2, and only afterwards testing a perceived pattern on benchmarks.

How do we try to explain the behavior of models, once we know there's a discrepancy we want to explain? Often we attempt to look at the qualitative differences between tasks a model is good or bad at, and come up with hypotheses for what the model is failing to do when it performs poorly, e.g., across different finetuning tasks \cite{li2021quantifying}. While useful for coming up with hypotheses, using benchmarks as evidence of behavior requires care, because it is often unclear what a given benchmark is actually testing. \citet{rohrbach2018object} show that image captioning systems hallucinate objects not present in the scene, and are unintentionally \textit{rewarded} for doing so by standard metrics, by capturing phrasing and $n$grams of reference captions better when hallucinating. \citet{liao2021we} describe a detailed framework for assessing benchmark validity and note the complexity of ensuring benchmarks test what we would like them to.
Thus, since we often do not know precisely what behavior benchmarks test, they might indicate what contexts to examine the \newmodel in, but not precisely what it does.

\subsection{Can benchmarks characterize behavior?}
\label{ssec:characterize}

\begin{CJK*}{UTF8}{gbsn}
Consider standardized tests for humans—such as the SAT \cite{national2008report} or the NCEE \cite{baike}—while the debate about how much these tests tell us is heated, there is little resistance to the statement: \textit{test scores do not fully describe human behavior}, even within the subjects they test such as mathematics and biology.
\end{CJK*}

Performance data about a bicycle is not sufficient to reverse-engineer its gear system. Even with perfectly valid benchmarks, 
the subspace of benchmark performance is not descriptive enough to characterize behavior. As we greatly increase the number of benchmarks, we are left with the problem of determining precisely how benchmarks overlap and differ in a way that characterizes behavior (Figure~\ref{fig:hourglass}). Because the space of benchmarks is limited, as we test for human-desirable skills and human-interpretable pitfalls, discovering novel behavior in non-human systems is difficult.

Measuring systems only for their expected purposes makes it difficult to disentangle component behaviors that allow models to produce the desired or undesired outputs, as failure under distribution shift often reveals. For example, neural machine translation often outputs completely irrelevant translations under domain shift \cite{wang2020exposure, muller2020domain}. This is exacerbated by the fact that most generative models are not trained with a precise purpose in mind.

Imagine testing whether an \lm can summarize an article. 
In order to summarize an article a requisite skill required by models is \textit{copying}, because novel entities are constantly appearing, but need to be referenced in the summary. \citet{see2017get} add a copying mechanism to an RNN in order to improve its copying ability for summarization.
If we were to only look at performance on summarization, we would be unlikely to notice whether copying was happening or not directly—only whether performance is hitting certain desired levels.

Benchmarks are, by necessity, scoped to certain contexts that are presumed to test for certain behaviors—but they do not directly tell us what patterns the model is exploiting to solve the task, as \citet{liao2021we} point out. This was a hard-learned lesson in many benchmarks, such as when it was discovered that SNLI \cite{poliak2018hypothesis, gururangan2018annotation} 
could be solved with \textit{hypothesis-only} systems that only use a subset of the information that was supposed to be necessary to the task.

\subsection{Behaviors: building blocks for evaluation}
\label{ssec:building-blocks}

Benchmarks are still the best solution for coordinating cross-lab experimental \textit{comparisons}, and we expect them to continue to be useful in that respect indefinitely. However, to answer ``\textit{What} strategy is the \newmodel using for this task?'' and ``What failure modes should we expect?'' and “What else do we expect the \newmodel to be capable of?''
we cannot use benchmarks alone to guide where we inspect model behavior, nor as a means to define it.

Instead, we propose an increased focus on  behavior, because we believe the science of generative models is currently held back by insufficient understanding of \textit{what} models are doing in general, rather than \textit{how well} models perform on specific tasks. These are highly related to each other, and we can think of \textit{behaviors as building blocks for evaluation}. Consider the following thought experiment:
\begin{equation*}
  \tag{B}\label{te:schemas}
  \parbox{\dimexpr\linewidth-4em}{
    \strut
    A new \lm{} is released with many of the expected capabilities, such as basic arithmetic and basic translation, but another interesting behavior is noticed and hypothesized: when asked properly in natural language, the model can stegonographically encode complex hidden messages while completing other tasks.
    \strut
  }
\end{equation*}
When this \lm{} is released it is unlikely there are any benchmarks that test this particular capability.
While we could design a specific benchmark for this behavior, this would be somewhat counter-productive: what we really care about is the \textit{Cartesian product} of this behavior and other tasks that we were already testing. In this sense, behaviors are the building blocks for benchmarks.

As \citet{chang2023language} point out in their survey of behaviors, researchers are often surprised by the outputs of the models they work with; it should not surprise us that we cannot premeditate benchmarks to capture behavior when modeling improvements have outpaced our ability to be exposed to generated data. One way to be more nimble to new behaviors, is to directly measure behaviors we expect \cite{jain2023bring}, flagging unexpected combinations for inspection.

On the surface, it might seem that naming behavioral categories such as ``copying'' or ``in-context learning'' is just as liable to obsolescence as any other analysis. What should we do if the \newmodel does not exhibit these behaviors? We argue that this is a very unlikely scenario: as long as we are attempting to train models to mimic human understandable phenomena, there will be human perceivable patterns that we expect models to mimic as well.

\clearpage

\section{Generative Models \hspace{120pt} as a Complex Systems Science}
\label{sec:complex}

While the \newmodel (\S\ref{sec:newmodel}) is a thought experiment, it is representative of many facets of research regarding generative models today; suddenly, focus has shifted to searching for \textit{emergent behaviors} in large and often inscrutable models.
Larger pretrained models continue to be trained and continue to perform better \interalia{openai2022chatgpt, OpenAI2023gpt4, Pichai2023-hp}. While efforts to release models \citep{Touvron2023-ml, falcon40b, stablelm} and involve more researchers in model training \citep{bigscience2022bigscience} can increase transparency and provide more information, it is well beyond the resources of the vast majority of labs to train. Efficiency breakthrough are likely to be exploited to further increase model size and feed into the same problem they were meant to solve.

\begin{figure}[t]
    \centering

    \includegraphics[width=\linewidth]{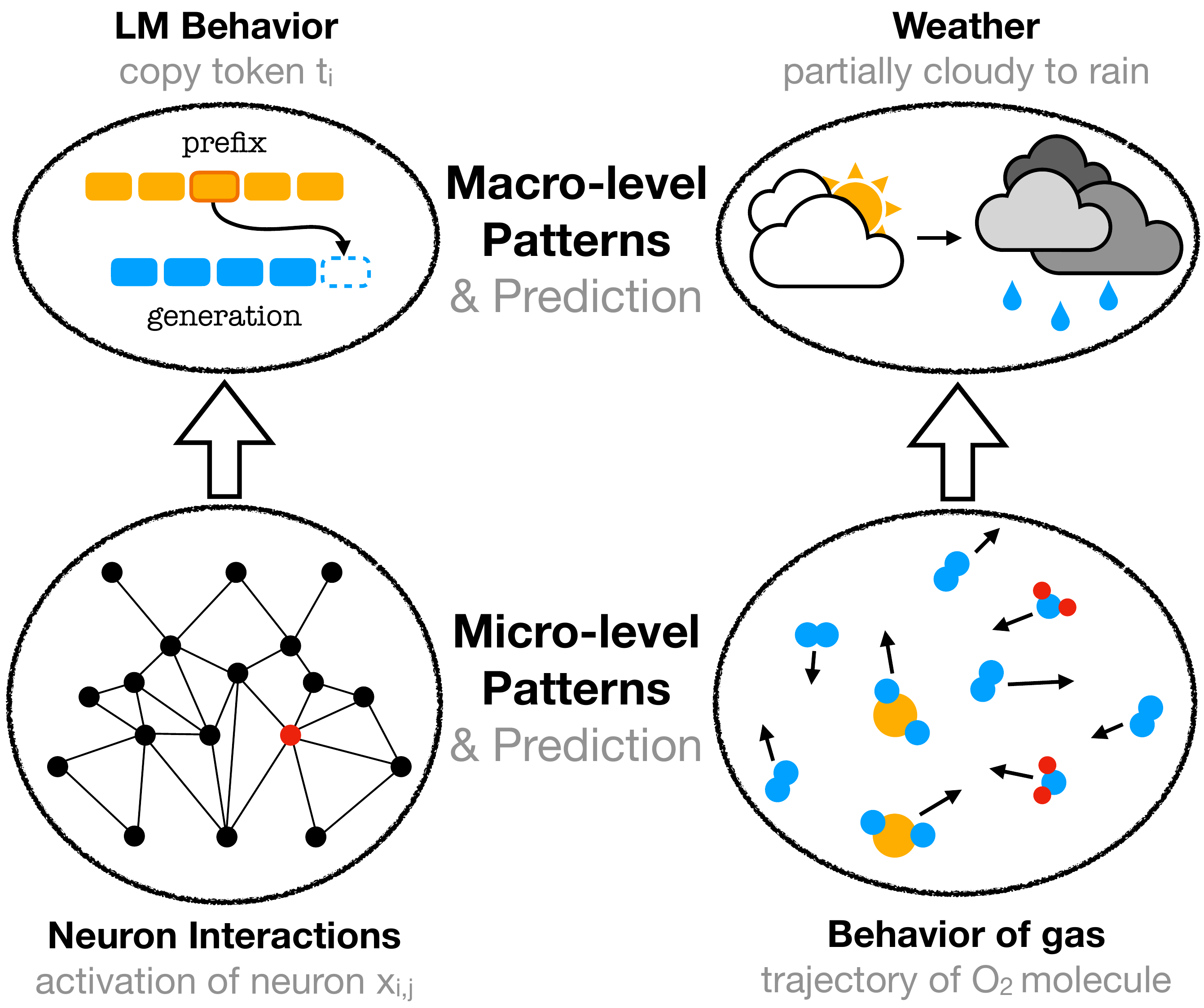}
    
    \caption{Complex systems are characterized by two or more levels of regularity: a micro-level in which local interactions are at least partially predictable and a macro-level in which many local interactions collectively exhibit recognizable patterns. 
    \textit{Emergence} describes how macro-level regularity is hard to predict in advance from comparatively well-understood micro-level dynamics.}
    \label{fig:micro-to-macro}
\end{figure}

Thus, it seems likely that training and re-training models is no longer the path towards understanding them for the vast majority of researchers. In many fields the creation of what it studies is impossible, from biology to astronomy. Many of these fields are \textit{complex systems sciences}, in that they focus on the question illustrated in Figure~\ref{fig:micro-to-macro}: how do the macro-level behaviors we observe (life, black holes, etc.) arise from the micro-level units we understand better (chemicals, regular matter, etc.)? 

In other words, we suggest studying \textit{generative models themselves not just generative modeling}.

\subsection{What is a complex system?}
\label{ssec:what-is} 

\citet{newman2011resource} establishes a working definition:
\begin{displayquote}
    {[A]} system composed of many interacting parts, such that the collective behavior of those parts together is more than the sum of their individual behaviors. The collective behaviors are...“emergent” behaviors, and a complex system can thus be said to be a system of interacting parts that displays emergent behavior.
\end{displayquote}
Recently, interest in emergent behavior has grown in NLP \cite[][\textit{inter alia}]{bubeck2023sparks, wei2022emergent, teehan2022emergent, manning2020emergent}, though it is usually defined, in terms of scaling over model parameters, dataset size, or computational power. We rely on a much simpler definition:
\newpage
\begin{displayquote}
    Emergent behaviors are system level behaviors that are hard to predict from the dynamics of lower level subcomponents.
\end{displayquote}
For instance, the ocean is a complex system. We can understand many properties of individual water molecules, e.g., \ce{H2O} has a partial positive and negative charge in certain places due its composition, but the aggregate properties of \textit{water} as a collective whole exhibits predictable properties such as waves. It is difficult to predict the properties of water from \ce{H2O} because ``the interactions of interest are non-linear...[yielding] levels of organization and hierarchies—selected aggregates at one level become ‘building blocks’ for emergent properties at a higher level, as when \ce{H2O} molecules become building blocks for water.'' \cite{holland2014complexity}

Similarly, we understand the basic mechanical properties of \lm{}s at the neuronal level, e.g., we have a perfect understanding of how to predict what any individual neuron will do given arbitrary inputs (by construction), but we also notice patterns at the level of \textit{model behavior}, e.g., the emergent copying behavior, which is observed in both Transformer models \cite{al2019character, khandelwal2019sample} and LSTM models \cite{khandelwal2018sharp}. In the face of new behavior that a model such as the \newmodel might exhibit, we would be even less certain of how lower-level system components add up to observed responses.

\subsection{Emergent behaviors in \lm{}s are discovered, not designed}
\label{ssec:design-vs-discover}

Neural architectural elements (e.g. position embeddings) and training methods (e.g. masking strategies) deeply affect the resulting model but do not fully explain behavior. We often fail to create the behavior we attempted to engineer into an architecture \textit{and} 
discover new, unintentional behavior.

Many architectures have been designed to make use of longer context \interalia{yu2023megabyte, beltagy2020longformer, child2019generating}, but evidence suggests that these models often do not make use of the long-term dependencies that they intended to capture \cite{Liu2023-qo, sun2021long, press2021shortformer}. 
Inversely, BERT was shown to capture much of the functionality of a knowledge base without task-specific training \cite{petroni2019language}.

To illustrate the difference between \textit{designing} and \textit{discovering} behavior, let us return to our running example of the copying behavior, where models produce a span that was in their input. A classic example of designing behavior is 
pointer-generator models~\cite{see2017get},
in which a specific, discrete mechanism was added to encourage a certain behavior: copying. Transformers, on the other hand, were designed such that computation at a given time-step could \textit{attend} to any previous time-step that was included in the context window. This intentionally removed the recurrence in architectures such as LSTMs  \cite{hochreiter1997long} and GRUs \cite{cho2014properties} in order to increase efficiency on highly parallelizable hardware such as GPUs and TPUs. A side-effect of this change was the emergent behavior of \textit{copying} that arises directly from the Transformer architecture trained as a language model \cite{al2019character}.

Instead of directly designing models for these purposes, we are now in the position of training general models with different structure and actively probing them for behavior. 
Using various data and masking strategies has produced models that can be controlled through different metadata \cite{keskar2019ctrl, zellers2019defending, aghajanyan2022htlm}, while instruction-tuning has shown that pretrained \lm{}s can be finetuned for control \interalia{mishra2022cross, chung2022scaling, ouyang2022training} often with very limited data \cite{zhou2023lima, dettmers2023qlora, alpaca}.

This \textit{discovery} process focuses on giving the model access to certain kinds of correlations, and then inspecting what model behavior emerges.

\subsection{Neuronal explanations are limited by \hspace{100pt} our understanding of behavior}
\label{ssec:neuronal}

It is difficult to explain how or why \lm{}s produce their outputs without having a good description of \textit{what} they do. Explaining behavior bottom-up, 
requires an understanding of what behaviors we are trying to explain. \citet{mittal2018emergent} note:
\begin{displayquote}
    an emergent property of a system is usually discovered at the macro-level of the behavior of the system and cannot be immediately traced back to the specifications of the components, whose interplay produce this emergence.
\end{displayquote}
This is the situation we find ourselves in with regards to large, pretrained models. We cannot, in general, predict how structure will form. While we can engineer systems with the hope of producing certain kinds of behavior, e.g., training on multimodal data to produce models that can draw inferences in ways that integrate paired text and images, this often does not produce the desired results \cite{ilharco2021probing, parcalabescu2021seeing}.

Bottom-up investigation can reveal key properties of emergent organization within \lm{}s, e.g. BERT replicates features of the classical NLP pipeline \cite{tenney2019bert}. But when anomalous behavior is discovered, e.g., the DALL•E 2 hypothesized ``hidden vocabulary'' of invented words that correspond to specific image categories \cite{daras2022discovering}, it is difficult to investigate them with bottom-up tools until we reach a better understanding of what triggers them, what their scope is, etc. There have been attempts to reject the hidden vocabulary hypothesis \cite{Hilton2022-ej}, but it is a very difficult hypothesis to rebut from first principles: what tests reject the hypothesis ``DALL•E 2 has a hidden set of vocabulary with clear and consistent meaning'' rather than  ``this specific mapping from the vocabulary to features isn't correct''?

This is similar to trying to research organic chemistry without knowledge of biology:  it is certainly not impossible, but without high-level guides to the kind of structure one is expecting, the search space is huge and it is difficult to know where to look. Our lack of a behavioral \goal hampers research into internal structure, especially in models that break current assumptions such as the \newmodel, as it is significantly more challenging to probe for structure without knowing what patterns in the outputs hint at the presence of structure. 

\newpage

\clearpage

\subsection{Access is not a silver bullet}
\label{ssec:access}

Consider the following thought experiment:
\begin{equation*}
  \tag{C}\label{te:release}
  \parbox{\dimexpr\linewidth-4em}{
    \strut
    Tomorrow, all industry labs publicly \\ release all of their pretrained models
    \strut
  }
\end{equation*}
Despite the fact that this would doubtlessly help us understand the basic properties of a given model such as ChatGPT, e.g., how large it is, we would still have significant obstacles on the way to explaining why ChatGPT is capable of writing short stories for almost any given prompt.

Indeed, the problem with answering the question of ``How can a language model write a story?'' has much less to do with language models and much more to do with the fact that we are currently incapable of answering the question ``How can $x$ write a short story?'' for any value of $x$. We find ourselves in the strange position of being able to train models we do not fully understand \textit{for tasks we do not fully understand or anticipate in advance}.

The key to answering this question is to ask: what kind of explanation would satisfy us? For instance, when it comes to \lm{}s, one explanation is that models are simply reconstructing long sequences from the training set and stitching them together. While a significant amount of memorization is taking place \interalia{mccoy2023much, carlini2023quantifying, lee2022deduplicating} models appear to be able to generate data that is not a trivial recombination of the training data \cite{bubeck2023sparks, tirumala2022memorization, olsson2022icl}.

The goal, then, should be to build up the case for a reasonable hypothesis that explain the breadth, depth, and (most importantly) mistakes models make when executing a complex task. However, we do not want a new explanation for every new task, which is precisely why we argue for the formalization and study of \textit{behaviors} that describe the underlying strategies of models.

While model access would not directly solve these problems, we \textit{do} believe that open-source models are a necessary prerequisite to this research program, for reasons outlined in \S\ref{ssec:opensource}.

\subsection{(Generated) data represents behavior}
\label{ssec:maac}

\begin{figure}[t]
    \centering

    \includegraphics[width=\linewidth]{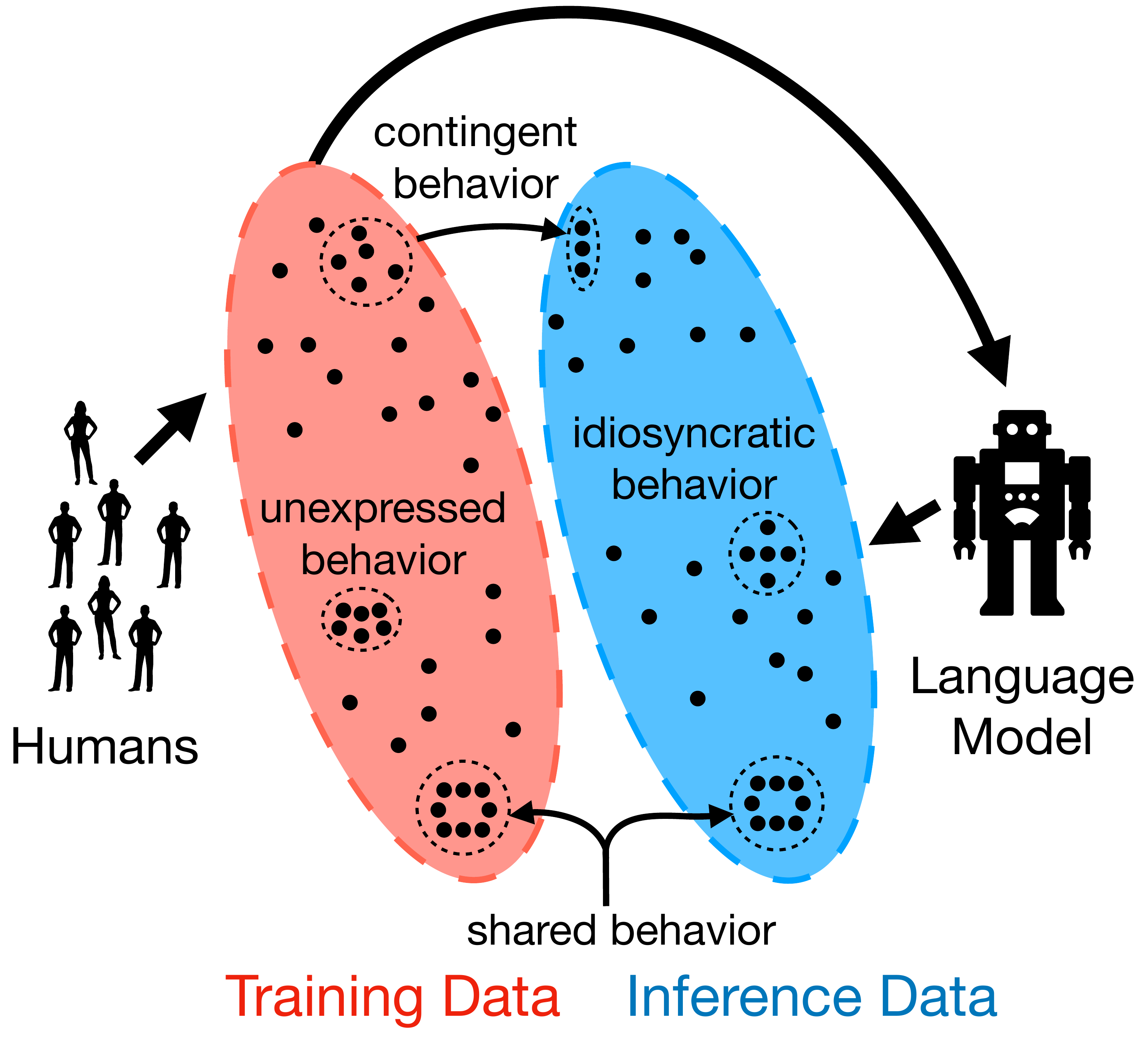}

    \caption{Generative language models are trained to capture the distribution of training data, then exhibit behavior in model outputs, i.e., \textit{inference data}.
    See \S\ref{ssec:maac} for examples of the different behavioral mappings.}
    \label{fig:maac}
\end{figure}

Behavior in large pretrained models is nothing more than the answer to the question ``How can we characterize the distribution of data this model generates?'' Aspects of the training data such as the presence of multiple languages \cite{blevins2022language, lin2021few} or the number of repeated documents \cite{kandpal2022deduplicating, lee2022deduplicating} in the training set have been shown to be explanatory of zero-shot translation abilities and model tendency to leak training data, respectively. 

Figure~\ref{fig:maac} visualizes what kind of behavioral mappings we can explore with data-based explanations. \textit{Shared behavior}—patterns that are found in both the training and inference data (the outputs of the model)—are the simplest to search for, because they only require finding a specific behavior in the training or inference data and then looking for it in the other. For instance, the prompting behavior discovered in GPT-2 that causes summaries to be generated when ``TL;DR'' is placed after an article is an example of shared behavior. Idiosyncratic behaviors describe behaviors that don't appear to be caused directly by the training data at all, e.g., zero-length translations in many large models \cite{stahlberg2022uncertainty, shi2020neural, stahlberg2019nmt}. Perhaps the most difficult to find behavioral mappings are those for which behavior in the corpus yields different behavior in the model, \textit{contingent behavior}, as is hypothesized to be the case for DALL-E 2's  ``hidden vocabulary'': nonsense words that appear to consistently lend certain meanings to produced images \cite{daras2022discovering}. Finally, unexpressed behavior is observed in the training data, but not in the inference data, such as long-term consistency in story telling \cite{Xie2023-qp, see2019massively} that models have yet to properly mimic for very long documents.

\section{A Different Kind of Complex System}
\label{sec:diff}

 One reasonable worry is that taking on the complex systems lens will be fruitless because studying complex systems is a very difficult task, and we are not equipped to tackle such a hard problem.

In fact, compared to other complex systems, such as the brain, understanding current generative models is an immensely \textit{easier} challenge, and can help us develop tools for the future. Turning our attention to ``What, precisely, do language models do?'' over ``What is the best recipe for training large models?'' we can take full advantage of the \textit{complete simulability} of generative models. In the long run, it seems it will become more difficult to address the latter question coherently without better answers to the former.

\subsection{Two kinds of complex systems simulations}
\label{ssec:two-kinds}

\begin{displayquote}
    Complex systems theory is divided between two basic approaches. The first involves the creation and study of simplified mathematical models that, while they may not mimic the behavior of real systems exactly, try to abstract the most important qualitative elements into a solvable framework from which we can gain scientific insight...The second approach is to create more comprehensive and realistic models, usually in the form of computer simulations, which represent the interacting parts of a complex system, often down to minute details, and then to watch and measure the emergent behaviors that appear. \cite{newman2011resource}
\end{displayquote}

At first glance, generative models can largely be described as the second complex systems approach: we train models to capture properties of the natural distribution of human media, such as internet text, images, videos, etc., and then attempt to study the emergent effects. 
Yet, this would be a mischaracterization of what, e.g., language models do: we do not expect that language models learn language the way a human does nor create languages the way the human species did.\footnote{Though this certainly describes facets of certain subfields such as emergent communication \cite{lazaridou2020emergent}, with recent work taking advantage of pretrained models \cite{Steinert-Threlkeld2022-mn}, and developmentally plausible pretraining such as the recent BabyLM challenge \cite{warstadt2023call}.}

Instead, the triumphs of generative models are the result of emergent behavior within computational models trained to predict very general objectives. Many have been surprised that by learning on massively more data from a given medium than a human is ever exposed to, generative models can learn uncannily human patterns from simple, passive word prediction, denoising objectives, etc. 

Generative models are certainly a computational simulation, but they are a simulation of an entire medium rather than a singular process we have isolated. We suggest thinking of generative models as a different kind of complex system, where \textit{underlying patterns of a medium are learned by a model through optimization, and we then look for those patterns within the model.} Below we list ways in which this discovery process is made easier because our system of interest is the computational model itself, rather than a naturally arising system.

\clearpage

\subsection{Generative Models: \hspace{120pt} the easiest complex system to study }
\label{ssec:advantages}

\begin{enumerate}[label=\textnormal{(\arabic*)}]
    \item \label{adv-state} \textbf{Perfect fidelity state encoding} Because neural networks are formal mathematical models, represented by code and parameters, there is zero \textit{necessary ambiguity} in our representations. Imperfect data archiving and complex code bases often make it difficult to perfectly recover the formal model, but with sufficient effort, it is possible to store every bit of information about the state of the model at every computation step. We cannot track every neuron in a participant's brain at every moment due to the limited nature of our measurement instruments, but we can perfectly record the state of an \lm{} in order to look for and verify emergent behavior, without influencing the system as we note in Advantage~\ref{adv-observer}.
    
    \item \label{adv-dynamics} \textbf{Complete theory of low-level dynamics} While Advantage~\ref{adv-state} establishes that we can perfectly store the state of a generative model at any given moment in time, Advantage~\ref{adv-dynamics} notes that we have a perfect, mechanistic, and deterministic understanding of how one state of a model evolves into the next, unlike in physical experiments. Artificial neural networks do not need to be simulated, they are defined in a medium for simulation: executable code. Unlike in physics, where centuries of research have been spent chasing the bottom of the chain of causation, we \textit{begin} with the base-level causal structure of the model. This does not follow directly from (1). It is possible to imagine a scenario where every static state is recordable, but where the rules that govern the changes in states are hard to discover, e.g., the problem of learning video game dynamics from pixels \cite{hafner2019learning}. In practice, nondeterminism exists in certain fast  computations \cite{morin2020non}, but this can be removed at the cost of speed.
    
    \item \label{adv-repeat} \textbf{Exact repeatability} Directly entailed by (1) and (2) is the fact that experiments can be repeated \textit{exactly}. An algorithm that uses randomness to generate text, may generate different text on a second run, but as long as the probabilities of different tokens are recorded the likelihood of that text (and of alternative branching paths) can be verified to be exactly the same. A psychologist who conducts a study twice will almost never get results that are exactly the same, simply because sample differences and unmeasured variables have to be accounted for. We distinguish repeatability from the broader notion of \textit{replicability}, which also includes replicating a study to the level of detail described by the authors, leaving room for both human and systematic error. With proper code, data, and model releases, many generative model experiments are exactly repeatable, allowing us to reach for a much higher standard for replicability.
    
    \item \label{adv-perturb} \textbf{Ease of perturbation}
    We also have a complete description of \textit{all possible models} given a certain setup, e.g., all possible combinations of weights for a given architecture. Combining this with Advantage~\ref{adv-dynamics}, we can perturb a model of interest, and play out experiments with this new model \textit{without destroying the original model}. Contrast this with studying human language production, for which most perturbations of the human brain are both unethical and illegal, partially because humans cannot be unperturbed. This allows for extremely targeted experiments, e.g., finding which weights in a network control a certain decision boundary.
    \item \label{adv-observer} \textbf{No observer effects}—a classic problem in many complex systems is that by attempting to make a measurement one changes the value being measured, e.g., Clever Hans a horse that could allegedly play chess, but was simply reading the audience's reaction to possible moves \cite{prinz2006messung}.
    In contrast, generative models do not distinguish between the same input given for different reasons or with different expectations by the experimenter.
    The caveat is that experimenters still control the input distributions to experiments allowing for systematic bias that accidentally leaks \textit{ experimenter expectations} \cite{rosenthal1976experimenter} to the model, as past research has consistently shown \interalia{mccoy-etal-2019-right, poliak2018hypothesis, gururangan2018annotation}. 
    We must be careful about ``tells'' \cite{caro2003caro}: stylistic and semantic artifacts that make it into the data which can give the model information the experimenter assume it does not have access to. Yet, the guarantee that the specific observer will not change the result is strong.

\end{enumerate}

Advantages \ref{adv-state} and \ref{adv-dynamics} allow us to completely remove any worry about hidden variables that may explain effects we attempted to explain through other means. \ref{adv-repeat} and \ref{adv-perturb},  allow us to experiment freely, knowing that experiments and models that have been properly recorded are recoverable, leaving us free to perturb and explore the local neighborhood of similar models and setups. \ref{adv-observer} partially relieves us of the fear of influencing the outcome through our means of observation, a key issue in many experiments involving language.

Another advantage, that does not apply to every generative model, deserves an honorable mention:

\begin{enumerate}[label=(6*)]
        \item \label{adv-human}\textbf{(Some) generative models exclusively output human understandable media} Many complex systems, such as cities and brains, produce human understandable media as some percentage of their output. Many generative models produce human understandable media as their \textit{only} output, an enormous advantage for two reasons. First, humans are better suited to positing patterns in human understandable media than, say, subatomic particles. Second, \textit{the uncanny valley effect} \cite{Mori2012-ew} allows us to see when patterns are ``almost correct but not quite'' much more easily in human-related artifacts. While we sometimes finetune models to produce outputs that are no longer human understandable, by and large current generative models operate entirely within human media—and we believe there is much that can be learned from this that will transfer over to generative models of other media.
\end{enumerate}

Advantage \ref{adv-human} is very special. By allowing us to take advantage of our intuitive understanding of media, it becomes easier to seek out the ways generated media diverges from the natural human media we are steeped in from infancy. Indeed, most named behaviors are failure modes, e.g. degeneration \cite{holtzman2020curious} or empty translations \cite{stahlberg2019nmt}).

Organic chemistry has given a great deal to biology, but is very much indebted to it as well.
Our hope is that we can take inspiration from these other complex system sciences to start taking the problem of understanding \textit{behavior} seriously, as a distinct abstraction that needs to be decomposed and theorized, while putting our enormous advantages to good use.

\subsection{The necessity of open-source models}
\label{ssec:opensource}

Most of these advantages rely on stable access to a consistent representation of a model, which is difficult to guarantee via a proprietary API.

\begin{itemize}[]
    \item[\cancel{(1)}] \textbf{Perfect fidelity state encoding} It is difficult to work with or guarantee saved state is persistent and untampered without direct access to said state. Even cryptographically signed state can be tampered \textit{after} re-submission to an API for use there, making guarantees moot.
    
    \item[\cancel{(2)}] \textbf{Complete theory of low-level dynamics} 
    With only imperfect knowledge of an underlying model, researchers must make assumptions about low-level dynamics in a model that may only partially be true, or possibly even completely false.
    
    \item[\cancel{(3)}] \textbf{Exact repeatability} In practice, it is impossible to guarantee that an API will not drift over time, something observed with even the apparent attempts at stable APIs in recent years \cite{Deng_undated-cl}.
    
    \item[\cancel{(4)}] \textbf{Ease of perturbation} It is normally impossible to perturb a model through an API, though some APIs allow for finetuning and special versions of models. However, the real issue is that it is impossible to ensure that such perturbations do precisely what they are claiming to do to the model, without access to the model or even the model architecture in most cases.
    
    \item[\cancel{(5)}] \textbf{No observer effects} Sadly, even though this is one of the greatest advantages of generative models, it is the one most destroyed by using models via APIs: companies consistently, and often silently, fix undesired (from the company's perspective) behaviors in models \cite{Wilson_undated-xd, Eliacik_undated-te, Kiho_undated-ov} so that testing a certain hypothesis tends to influence future tests.

    \item[\cancel{(6*)}] \textbf{(Some) generative models exclusively output human understandable media} Without complete access to a model it is impossible to know if it doesn't have other outputs (or inputs) that would help explain the model's behavior more fully.

\end{itemize}

In short, without access to open-source models, these advantages are largely moot. However, the community has seen a consistent open-source releases of better generative models in many different media \cite{rombach2021highresolution, Le2023-uv, VideoFusion}. There is unquestionably lag in the capabilities between proprietary and open-source models, and this is out of necessity: open-source cannot outpace private industry when private industry controls most of the training resources and can build on top of anything open-source does. But the fact that open source often lags only a year or two behind in terms of capabilities, and the fact that private labs are often incentivized to open-source models as a recruiting and market strategy, suggests that open-source will continue to be a wellspring of fascinating generative models to study. Indeed, if all progress stopped now, we believe it would be decades before we finished cataloging all of the generalizable behavioral principles with the hundreds of large generative models that have already been released; perhaps our successes would encourage future open-source releases.
\section{Conclusion}
\label{sec:conclusion}

How should we study models of data, when we don't fully understand the models or the data? We should study them first by asking \textit{what} models do, before attempting the more complicated \textit{how} and the bottomless question of \textit{why}?

In this paper, we presented a thought experiment: the \newmodel, a model that would be impossible to study with many of the techniques we use to understand Transformer models today. 

We argue that focusing on what \textit{behaviors} explain its performance across tasks will lead-us to a deeper understanding of generative models' tendencies and guide bottom-up mechanistic explanation, as well as forming building blocks for evaluations.

We discuss how generative models are well captured by the definition of a complex system, due to the emergent behaviors they exhibit. This separates generative models from traditional machine learning, where models often served as explanations via behaviors that were architected directly into them. This opens up the need for \textit{metamodels} that help us predict regularities in generative model outputs in order to understand them better.

While the prospect of studying models we do not have a clear understanding of is daunting, we highlight  advantages that generative models have over naturally arising complex systems. These advantages, however, require open-source models as a prerequisite, a point we emphasize as a necessity for conducting replicable science.

\section{Limitations}

We present one perspective on the kind of science NLP is becoming, and how we can leverage the complex systems lens in order to better explore the phenomena we find ourselves faced with: generative models we do not fully understand. We cite evidence from NLP publications, blog posts, and other media, but this necessarily does not capture the totality of perspectives. 

Indeed, we purposefully avoid attempting any sort of survey of these issues, as this would involve citing thousands of papers and be a very unwieldy object. Instead, we attempt to form an argument as economically as possible, attempting to put forth a new set of goals and principles for how to study generative models given current progress.

We make comparisons with other sciences and cite sources from those sciences where appropriate, but are extremely limited in expressing many equally relevant connections and in fully exploring the connections we do mention. There is an enormous amount related to sister fields (e.g., cognitive science, linguistics, etc.), other sciences that study complex systems (e.g., chemistry, biology, etc.), and regarding more meta-science issues (e.g., complex systems theory, chaos theory, etc.) that we could not cover, and we do not in any way attempt to—giving a complete account of these connections is simply beyond the reach of any one work.

Finally, parts of our assessment is necessarily subjective. We attempt to lay out the evidence as we see it, tracing the connections we drew in order to describe a style of research that we believe is necessary to face the current challenges of our field. This seems especially pertinent in a time when most researchers cannot train large generative models from scratch, but are excited to contribute to their study. With evidence drawn from the literature, we describe the current research space as we perceive it, and our vision for where it might go. Our hope is that this will add to a discussion on what the study of generative models currently is and what we, as a community, would like it to become.
\section*{Acknowledgements}
We thank Julian Michael, Dallas Card, Jared Moore,  Daniel Fried, Gabriel Ilharco, Tim Dettmers, Ian Magnusson, Alisa Liu, and Kaj Bostrom for their insightful discussions and feedback.

\bibliography{custom}
\bibliographystyle{acl_natbib}

\end{document}